%% file: main.tex
\RequirePackage[svgnames]{xcolor}

\documentclass[11pt,letterpaper]{mystyle}

\usepackage[all]{hypcap}
\usepackage[comma,authoryear,compress]{natbib}
\hypersetup{
    colorlinks = true,
    linkcolor = {MINDSPlum},
    citecolor = {AcademicTeal},
    urlcolor = {AcademicTeal},
}

\usepackage{algorithm}
\usepackage{algpseudocode}
\usepackage{mathtools}
\usepackage{mathrsfs}
\usepackage{nicefrac}
\usepackage{dsfont}
\usepackage{cleveref}
\crefname{appendix}{Appendix}{Appendices}
\Crefname{appendix}{Appendix}{Appendices}
\usepackage{bxcoloremoji}
\usepackage{wrapfig}
\usepackage{stackengine}
\usepackage{adjustbox}
\usepackage{rotating}
\usepackage{makecell}
\usepackage{array}
\usepackage{multirow}
\usepackage{longtable}
\usepackage{subcaption}
\usepackage{bigstrut}
\usepackage{float}
\usepackage{placeins}
\usepackage{booktabs}

\input{macro}

\newtcolorbox{AIbox}[2][]{aibox,title=#2,#1}

\title{LLM-Guided Heuristic Design from Simulation Traces: A Case Study in Dynamic Production and AGV Scheduling}

\runningtitle{LLM-Guided Heuristic Design from Simulation Traces: A Case Study in Dynamic Production and AGV Scheduling}

\author[1]{Jinbo Li}
\author[1]{Chuanhao Li}

\affil[1]{Department of Industrial Engineering, Tsinghua University}

\correspondingauthor{Chuanhao Li; Email \href{mailto:chuanhao-li@tsinghua.edu.cn}{chuanhao-li@tsinghua.edu.cn}}

\begin{document}

\input{sections/abstract}

\maketitle
\input{sections/introduction}
\input{sections/relatedwork}
\input{sections/method}
\input{sections/experiments}
\input{sections/discussions}
\input{sections/conclusion}


\clearpage
\bibliography{main}

\appendix
\input{sections/appendix}
\end{document}

%% file: macro.tex
\newcommand{\x}{\mathbf{x}}

\definecolor{blanchedalmond}{rgb}{1.0, 0.92, 0.8}
\definecolor{carmine}{rgb}{0.59, 0.0, 0.09}
\definecolor{lightblue}{rgb}{0.22,0.45,0.70}%

\makeatletter
\def\Ddots{\mathinner{\mkern1mu\raise\p@
\vbox{\kern7\p@\hbox{.}}\mkern2mu
\raise4\p@\hbox{.}\mkern2mu\raise7\p@\hbox{.}\mkern1mu}}
\makeatother

\definecolor{amaranth}{rgb}{0.9, 0.17, 0.31}
\definecolor{antiquebrass}{rgb}{0.8, 0.58, 0.46}
\definecolor{antiquefuchsia}{rgb}{0.57, 0.36, 0.51}
\definecolor{chromeyellow}{rgb}{0.31, 0.47, 0.26}

%% file: sections/abstract.tex
\begin{abstract}
Simulation-based optimization (SBO) evaluates executable policies under stochastic dynamics, but most methods treat the simulator as a black box: aggregate scores rank candidates without revealing why they fail or which policy logic should change. We present an LLM-guided heuristic design framework that uses repeated simulation for selection and event-level traces for diagnosis. Each incumbent is assessed through multiple replications, while replaying its lowest-scoring one produces a queryable trace. A manager agent formulates bottleneck hypotheses from this evidence, and editing agents implement parallel code-level revisions. After execution checks and repeated evaluation, best-so-far selection retains only improvements. LLM revision occurs between evaluation batches, while a fixed policy controls each simulation run.

We evaluate the framework in a discrete-event simulation of dynamic production and automated guided vehicle (AGV) scheduling. Across five independent optimization runs with Gemini-3.1-Pro, final mean scores averaged 77.51 on the simulator's 0-100 scale. In the highest-scoring run, trace-based diagnoses motivated proactive charging, distance-aware AGV assignment, and rebalanced dispatch priorities, raising the best-so-far mean score from 62.49 to 78.61. On 100 matched seeds, the best final policy outscored representative rolling-MILP, rule-based, and metaheuristic policies on every seed and retained its advantage under random faults without re-optimization. After separate re-optimization for a longer horizon and variable order interarrival times, the resulting policies again outscored all baselines. Ablations with two LLM backbones showed that removing either parallel candidate generation or trace-database access reduced final mean scores. These results show that simulation traces can guide targeted code-level policy improvement in complex simulation-based scheduling.

\vspace{2mm}


\end{abstract}

%% file: sections/introduction.tex
\vspace{-4mm}
\section{Introduction}
\label{sec:intro}

Designing and operating complex systems requires understanding how candidate decisions affect performance. Mathematical programming represents a decision problem through explicit variables, objectives, and constraints, often requiring simplifying assumptions for tractability. Simulation serves a different but complementary role: it evaluates specified decisions through a model of system evolution and can capture stochastic dynamics, resource interactions, and operational rules that may be difficult to retain in a mathematical programming formulation. Discrete-event simulation (DES), for example, represents system evolution through events that change the system state at discrete points in simulated time, such as arrivals, operation completions, and failures, making it well suited to complex systems with interacting components~\citep{law2015simulation,banks2010discrete}. This operational detail helps reveal how the effects of candidate decisions unfold through the modeled system over time. Simulation-based optimization (SBO) uses simulation as an evaluator in an iterative search procedure: candidate designs, parameter settings, or operational policies are repeatedly simulated, and the resulting metric values guide subsequent proposals~\citep{fu2002optimization,gosavi2015simulation,fu2015handbook}.

Recent advances in large language models (LLMs) create two complementary opportunities for SBO. Although developing and adapting DES models can be labor-intensive~\citep{tako2015modeldevelopment,robinson2004reuse}, emerging studies suggest that LLMs can help practitioners translate textual or formal specifications into executable models and modify simulation code as system requirements change~\citep{chen2026devsgen,schmitt2026llmdes,kramer2026automated}. Such assistance could lower the barriers to building and maintaining simulators and allow practitioners to devote more attention to verification and validation, thereby strengthening the modeling foundation for SBO. On the search side, LLM-based methods for program search and heuristic design, including Evolution of Heuristics, FunSearch, and ReEvo, represent candidate algorithms as executable code and combine LLM-generated variants with automated evaluation and evolutionary search~\citep{liu2024eoh,romera2024funsearch,ye2024reevo}. Together, these developments motivate coupling an existing simulator with LLM-guided search over executable decision policies.

One advantage of mathematical programming is that problem structure is made explicit in the formulation, allowing solvers to exploit relationships among decisions, constraints, and objectives during optimization. Many derivative-free and metaheuristic SBO methods, by contrast, treat the simulator as a black box: a search algorithm submits a candidate solution and receives estimates of objective and constraint values, while the modeled dynamics remain internal to the simulator~\citep{fu2002optimization,amaran2016simulation}. This separation allows complex dynamics to be evaluated without expressing them algebraically, but it provides limited guidance for candidate generation. Aggregate KPIs can rank alternatives, yet they do not reveal the operational mechanisms responsible for an outcome or indicate which condition, priority, or threshold in an executable policy should change. The LLM-based heuristic-design methods above likewise use automated performance evaluation to select executable candidates; ReEvo additionally conditions generation on textual reflections, but none uses event-level operational traces from a simulator as input to candidate generation~\citep{liu2024eoh,romera2024funsearch,ye2024reevo}. Simulators, however, can record process-level operational traces, hereafter termed simulation traces, including timestamped state transitions, resource states, queues, delays, command outcomes, and time-varying KPIs. They provide evidence about the operational mechanisms underlying aggregate outcomes. These feedback signals can therefore play complementary roles: mean scores support candidate selection, while simulation traces provide diagnostic evidence for more targeted candidate generation.

This paper presents a framework for LLM-guided heuristic design from simulation traces, in which each candidate is an executable decision policy. A manager agent uses the incumbent policy and trace evidence from a diagnostic replay associated with the lowest-scoring replication to formulate bottleneck hypotheses and propose multiple revision directions, which editing agents implement in parallel. Candidates that pass execution checks are evaluated over multiple stochastic replications; mean scores and best-so-far selection determine whether the iteration winner replaces the incumbent. LLM-guided revision occurs between evaluation batches, while each fixed policy makes operational decisions during a simulation run.

We evaluate the framework through a DES case study of dynamic production and automated guided vehicle (AGV) scheduling. Prior studies schedule machine operations and AGV movements jointly because operation completion times create transport requests, while AGV delivery times constrain the start of downstream operations~\citep{bilge1995time,ulusoy1997genetic,han2024dual}. In this case study, the executable policy assigns products awaiting release from a shared raw-material warehouse to one of three production lines, ranks transport tasks, dispatches line-specific AGVs, and controls proactive charging. Product routes and local workstation sequencing remain under simulator control. Dynamic orders, finite buffers, re-entrant processing, quality-induced rework, charging constraints, and optional equipment faults create interacting delays and bottlenecks that can be examined through simulation traces.

The main contributions of this paper are as follows:
\begin{itemize}[leftmargin=7mm,itemsep=1mm, topsep=0em]
    \item We propose a framework for LLM-guided heuristic design that uses mean scores for candidate selection and simulation traces to formulate bottleneck hypotheses and guide targeted revisions.
    \item We instantiate the framework as a manager--editor multi-agent loop that combines bottleneck analysis from simulation traces, parallel revision of policy code, execution checks, repeated simulation, and best-so-far selection.
    \item We evaluate the framework in a dynamic production and AGV scheduling case study against a rolling-MILP transport baseline, rule-based policies, and metaheuristic search over predefined rule combinations. We complement these comparisons with iteration-level analysis, tests under changed simulation settings and equipment faults, and component ablations.
\end{itemize}

%% file: sections/relatedwork.tex
\section{Related Work}
\label{sec:related_work}

Our work draws on three strands of research: work on production and AGV scheduling establishes the case-study context, SBO provides the evaluation-and-search paradigm, and LLM-based heuristic design provides mechanisms for revising executable policies.

\subsection{Dynamic Production and AGV Scheduling}

Production and transport decisions are coupled when AGVs move workpieces among processing resources. Job-shop scheduling provides a common abstraction for studying this interaction: each job comprises an ordered sequence of machine operations. When AGVs carry workpieces between machines, a downstream operation can begin only after the required transport is completed. Machine scheduling, transport timing, and vehicle assignment therefore become interdependent. Classical studies addressed this integrated problem through mathematical and time-window formulations, as well as genetic and hybrid metaheuristics~\citep{bilge1995time,ulusoy1997genetic,abdelmaguid2004hybrid,deroussi2008simple}. More recent work combines an MILP model for small instances with dual-population search using problem-specific encodings and decoders~\citep{han2024dual}.

Work on dynamic settings extends this literature to uncertainties and real-time decisions. \citet{zhang2013hybrid} study job-shop rescheduling under random arrivals and machine breakdowns, whereas \citet{umar2015hybrid} coordinate job scheduling and AGV routing in a dynamic flexible manufacturing system. Other work schedules machines and AGVs concurrently through real-time multi-agent negotiation~\citep{erol2012multiagent} or treats AGV dispatch, repositioning, and charging under randomly arriving transport requests as an online decision problem~\citep{singh2024dispatching}. Our case retains this production--transport dependency but optimizes product-to-line assignment and AGV task ranking, dispatch, and charging. Individual machine operations, product routes, and local workstation sequencing remain simulator-controlled. These policy decisions are evaluated in a dynamic DES, linking the case study to the broader SBO literature.

\subsection{Simulation-Based Optimization}

SBO couples an optimization method with a simulator, enabling candidate evaluation when the relevant input--output relationship is stochastic, discontinuous, unavailable in closed form, or difficult to represent in a tractable mathematical program~\citep{fu2002optimization,gosavi2015simulation,fu2015handbook,amaran2016simulation}. In manufacturing, DES has long supported system design, capacity analysis, and operational policy evaluation~\citep{law2015simulation,banks2010discrete,negahban2014simulation}. In stochastic combinatorial optimization, simheuristics combine metaheuristic search with simulation evaluation to handle uncertainty~\citep{juan2015review}. In a dynamic job-shop setting, \citet{wang2023data} combine simulation, machine learning, and evolutionary search to generate dispatching rules under random arrivals, stochastic processing times, and machine breakdowns.

Although a simulator may generate detailed trajectories, most SBO methods use it as a black-box evaluator. In a common formulation, the simulator maps decision variables to objective and constraint estimates~\citep{fu2002optimization,amaran2016simulation}. This supports candidate ranking but leaves simulation traces outside candidate generation, though they can provide diagnostic evidence about why a policy performs poorly and which parts of its logic may warrant revision.

\subsection{LLMs for Heuristic Design and Simulation Feedback}

LLM-based heuristic-design methods provide a mechanism for revising executable logic under automated evaluation. EoH jointly evolves heuristic concepts expressed in natural language and their executable implementations~\citep{liu2024eoh}. FunSearch searches over program fragments and has discovered useful mathematical constructions and heuristics~\citep{romera2024funsearch}. ReEvo uses LLMs as hyper-heuristic operators to evolve heuristics through reflective feedback~\citep{ye2024reevo}, while AlphaEvolve employs an evolutionary coding agent to edit programs across algorithmic and engineering tasks~\citep{novikov2025alphaevolve}. Although their architectures differ, these methods share a generate--evaluate--revise pattern in which executable candidates are tested and the results inform further code generation.

Some related methods make the feedback for code revision richer than a single final score. Eureka performs evolutionary optimization over reward-function code for reinforcement learning. Candidate reward programs are selected by the task fitness of the policies trained under them, while reward reflection summarizes reward-component and task-fitness histories collected during policy training to guide subsequent edits~\citep{ma2024eureka}. \citet{cheng2024trace} formulate Optimization with Trace Oracle (OPTO) for general computational workflows. Its oracle returns feedback on the output together with an execution trace represented as a computational dependency graph, and its OptoPrime optimizer uses an LLM to update heterogeneous workflow parameters such as prompts and code. Both methods move beyond scalar feedback, but Eureka uses reward-component and policy-training statistics, while OPTO records dependencies within a computational workflow rather than event-level operations in a DES. The present framework instead combines repeated scalar evaluation with a selected diagnostic trace to revise decision-policy code.

Within scheduling, LLMs have been used to generate heuristics for shop-scheduling benchmarks and to evolve dispatching rules for dynamic job shops~\citep{yu2026lsh,huang2024seevo}. MAEF coordinates multi-agent evolutionary search~\citep{wang2025maef}, while DSevolve evolves a portfolio of dispatching rules offline and selects among them online using look-ahead simulation~\citep{huang2026dsevolve}. More directly related to the present optimization loop, \citet{may2026actorcritic} combine DES evaluation with an actor LLM that generates dynamic-job-shop dispatching rules and a critic LLM that produces performance- and structure-based feedback for iterative refinement. The present work formalizes a different feedback interface: the mean score determines selection across repeated stochastic evaluations, while a selected, queryable diagnostic trace guides revisions to executable policy code.

LLMs have also been applied at other stages of simulation workflows. Some systems generate or adapt executable DES models from textual or structured specifications~\citep{chen2026devsgen,schmitt2026llmdes,kramer2026automated}. Others use LLMs to search simulation parameters~\citep{xia2024llm}, generate human-refined optimization formulations evaluated through DES~\citep{elbaz2026warehouse}, or make decisions within a running simulation~\citep{dehghani2026simgpt}. These studies target simulator construction, parameter search, formulation generation, or within-run decisions. By contrast, the present study assumes an existing, instrumented simulator and focuses on revising executable policy code between evaluation batches.

%% file: sections/method.tex
\section{Methodology}
\label{sec:method}

\subsection{Framework Overview}

Let $\Pi$ denote the set of executable decision policies. For one simulation run with policy $\pi\in\Pi$ and stochastic input $\xi$, the simulator returns a scalar score $S(\pi;\xi)$. The policy-design objective is
\begin{equation}
\pi^{\star}\in\arg\max_{\pi\in\Pi}J(\pi),
\qquad
J(\pi) \coloneqq \mathbb{E}_{\xi}\!\left[S(\pi;\xi)\right].
\label{eq:sbo_objective}
\end{equation}
Here, $\pi$ denotes the executable decision policy evaluated by the simulator. Fixed simulator settings and any evaluation context supplied to the policy are omitted from the notation.

The framework estimates $J(\pi)$ using $R$ independent simulation replications. With $S_r(\pi)=S(\pi;\xi_r)$ denoting the score from replication $r$, the sample mean is
\begin{equation}
\bar{S}_{R}(\pi)
=\frac{1}{R}\sum_{r=1}^{R}S_r(\pi).
\label{eq:aggregate_performance}
\end{equation}
A complete evaluation also returns trace-based diagnostic evidence, denoted by $\tau(\pi)$. It contains a queryable simulation trace and may also include derived summaries supplied to the revision mechanism. Depending on the application, the trace may be selected from one replication, aggregated across replications, or produced by an additional replay. We summarize the complete evaluation procedure as
\begin{equation}
\operatorname{Evaluate}_{R}(\pi)
=\left(
\bar{S}_{R}(\pi),
\tau(\pi)
\right).
\label{eq:evaluation_interface}
\end{equation}
The mean score $\bar{S}_R(\pi)$ ranks candidates and determines incumbent promotion, while $\tau(\pi)$ supplies trace-based evidence for revision. Both outputs depend on the realized stochastic inputs; the notation leaves this dependence implicit. No decomposition of $S$ is required.

Starting from an initial policy $\pi_0$, the framework alternates between LLM-guided revision and simulation evaluation. Let $\bar{S}_t=\bar{S}_R(\pi_t)$ and $\tau_t=\tau(\pi_t)$ denote the evaluation outputs retained with the incumbent $\pi_t$, i.e., the current best-so-far policy. At iteration $t$, the revision mechanism examines the incumbent and its retained feedback and produces $k_t$ executable candidates:
\begin{equation}
\left\{\pi_{t,j}\right\}_{j=1}^{k_t}
\leftarrow
\operatorname{Revise}_{t}\!\left(
\pi_t;
\bar{S}_t,
\tau_t
\right).
\label{eq:policy_revision}
\end{equation}
The revision mechanism may also use information retained from earlier iterations. Each candidate is evaluated using \Cref{eq:evaluation_interface}; write $\bar{S}_{t,j}=\bar{S}_R(\pi_{t,j})$ and $\tau_{t,j}=\tau(\pi_{t,j})$ for its outputs. The candidate with the highest mean score replaces the incumbent only if it improves on the incumbent's stored mean score; otherwise, the incumbent and its evaluation outputs are retained.

\Needspace{0.32\textheight}
\begin{figure}[t]
    \centering
    \includegraphics[width=\linewidth]{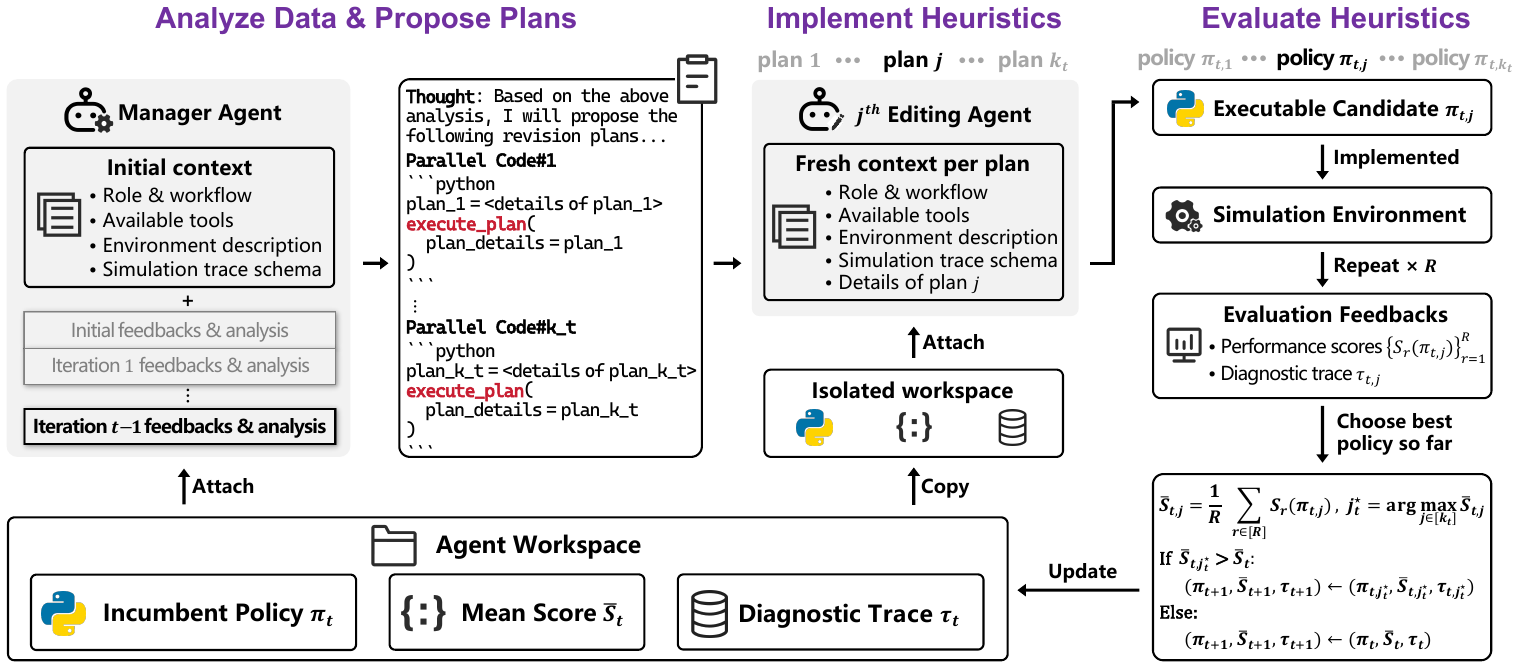}
    \caption{Overview of LLM-guided heuristic design from simulation traces. The incumbent policy and its retained feedback guide parallel code-level revisions. Each valid candidate is assessed using $R$ scoring replications, which provide a mean score for selection. Replaying the stochastic input from the lowest-scoring observed replication produces the diagnostic trace; its replay score is excluded from the mean score used for selection. The best improving candidate and its retained feedback become the next incumbent.}
    \label{fig:framework_overview}
\end{figure}

The framework is designed for problems in which executable policies can be evaluated repeatedly by a simulator, the simulator exposes queryable simulation traces, and candidate policies can be checked and run automatically. The next subsection instantiates this evaluation interface for the dynamic production and AGV scheduling case study.

\subsection{Simulation Environment and Trace Feedback}

The case-study simulator and diagnostic-replay procedure instantiate the evaluation interface in \Cref{eq:evaluation_interface}. The following paragraphs describe the simulated system, the decisions controlled by a policy, the score and trace records returned by the simulator, and the construction of the diagnostic trace.

\begin{figure}[H]
    \centering
    \includegraphics[width=\linewidth]{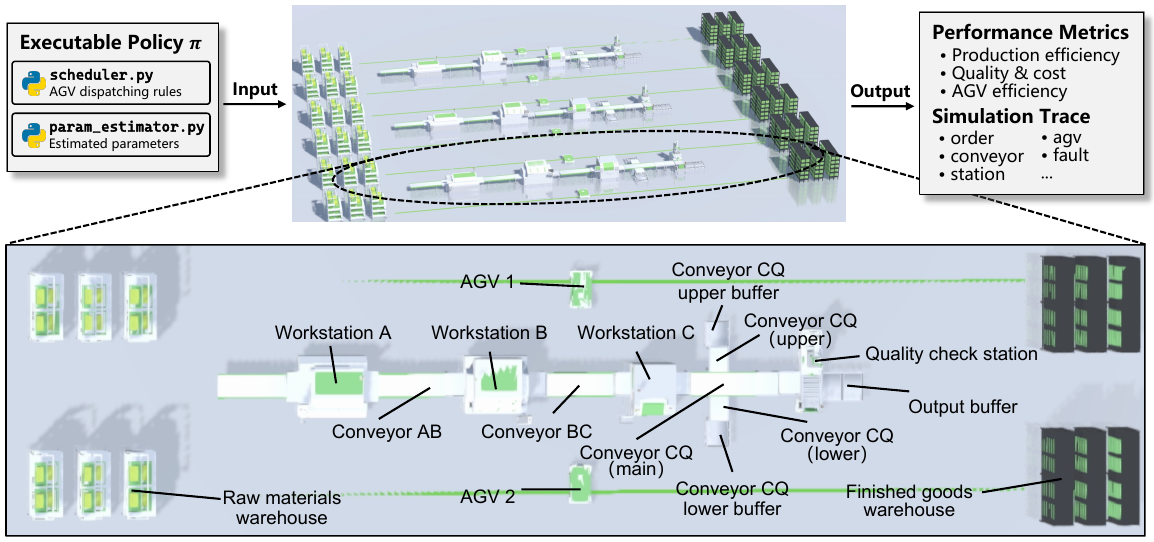}
    \caption{Simulation environment used as the feedback module in the dynamic production and AGV scheduling case study. The upper view shows the three-line factory and configurable dynamic-order and simulation-horizon settings; the enlarged view shows the principal production, storage, and transportation resources.}
    \label{fig:simulation_case}
\end{figure}

\textbf{Simulation system.}
The case study uses a three-line manufacturing DES model adapted from the FreezoneX 2025 smart-factory scheduling environment\footnote{\url{https://github.com/supcon-international/25-AdventureX-SUPCON-Hackathon}} and instrumented to return an aggregate score and expose event-level simulation traces to the optimization loop. As illustrated in \Cref{fig:simulation_case}, the system comprises workstations, conveyors, buffers, quality-inspection facilities, warehouses, and AGVs distributed across three production lines. The model also represents dynamic orders, rework, charging, and random faults. Detailed product flows are reported in \Cref{app:product_flows}.

\textbf{Policy representation and execution.}
Each policy comprises executable rules for production-line selection, transportation-task ranking, AGV assignment, and charging, together with optional parameter-estimation logic. At each decision point, these rules map the current system state to line-assignment, transport, and charging decisions. Product routes and local workstation sequencing remain under simulator control, while policy actions interact with conveyor transfers, quality rework, and buffer blocking.

\Needspace{0.62\textheight}

Before evaluation at iteration $t$, each candidate receives a fixed copy of the trace-database component of $\tau_t$ for pre-run parameter estimation. All $R$ scoring replications and the diagnostic replay for that candidate read the same snapshot, and parameters estimated from it remain fixed within each run. The scoring replications do not update the snapshot. Only the diagnostic replay writes the new trace retained for that candidate; if the candidate is promoted, this trace provides the snapshot copied for candidates at iteration $t+1$.

\Needspace{0.24\textheight}
\textbf{Score aggregation and performance metrics.}
In this case study, $S_r(\pi)$ is referred to as the total score. The simulator converts eight performance measures to normalized scores in $[0,100]$ and computes the total score as their fixed weighted sum. The weights sum to 100\%, so the total score also lies in $[0,100]$. For reporting and diagnostic context, the evaluator provides group-level KPI summaries for production efficiency, quality and cost, and AGV efficiency; the metric weights assigned to these groups sum to 40\%, 30\%, and 30\%, respectively. The summaries are included in $\tau(\pi)$ but are not separate objectives or selection criteria: candidate ranking and incumbent promotion use only the total score. The metric definitions and individual weights appear in \Cref{app:metrics}. This grouping is specific to the case study and is not required by the general framework.

\textbf{Simulation traces.}
The queryable trace in $\tau(\pi)$ is implemented as a database containing order events, resource-state changes, queue and buffer states, AGV tasks and charging events, fault and recovery events, and time-varying values of the underlying performance metrics. The manager queries relevant records or summaries rather than placing the complete database in the LLM context. The recording scheme appears in \Cref{app:trace_recording}.

\textbf{Diagnostic replay.}
Within each call to $\operatorname{Evaluate}_{R}$, the queryable trace is constructed by replaying the stochastic input associated with a lowest-scoring replication. Its replication index is selected as
\begin{equation}
r(\pi)
\in\arg\min_{1\le r\le R}S_r(\pi).
\label{eq:diagnostic_seed_selection}
\end{equation}
The policy is replayed with $\xi_{r(\pi)}$, and the resulting event history forms the queryable-trace component of $\tau(\pi)$. Because a simulation run is deterministic given the policy and the stochastic input, the replay reproduces the selected replication. A complete evaluation therefore comprises $R$ scoring replications and one diagnostic replay. The replay contributes diagnostic evidence; its score duplicates the already-counted $S_{r(\pi)}(\pi)$ and is therefore excluded from $\bar{S}_{R}(\pi)$. This choice focuses diagnosis on a poorly performing observed replication, while other applications may select or aggregate traces differently.

\subsection{LLM-Agent Optimization Loop}

We implement $\operatorname{Revise}_t$ using one manager agent and multiple editing agents, a role-specialized pattern used in general LLM multi-agent frameworks~\citep{hong2024metagpt,wu2024autogen}. The manager examines the incumbent policy, its mean score, retained diagnostic evidence, and prior search history, then proposes several distinct revision directions; the number of directions $k_t$ is chosen by the manager within a suggested range (\Cref{app:runtime_parameters}). In this case study, the diagnostic evidence combines the grouped KPI summaries with records and summaries queried from the trace database. Each editing agent implements one revision direction as an executable candidate policy. These diagnoses serve as working hypotheses for revision; simulation re-evaluation determines whether the resulting candidates are retained. \Cref{alg:framework} summarizes the procedure.

LLM agents revise policies only between evaluation batches; each candidate remains fixed during a simulation run. A candidate must complete the repeated scoring simulations and diagnostic replay in \Cref{eq:evaluation_interface} before it is eligible for selection. A loading or runtime failure triggers a bounded number of editing-agent repair attempts, with the complete evaluation restarted after each repair. A candidate that continues to fail is discarded.

\begin{algorithm}[h]
\caption{LLM-guided heuristic design from simulation traces}
\label{alg:framework}
\footnotesize
\begin{algorithmic}[1]
\Require Executable initial policy $\pi_0$, simulator, LLM-agent revision mechanism, maximum iterations $T$, number of scoring replications $R$, repair-attempt limit $N_{\mathrm{repair}}$, no-improvement patience $C_{\mathrm{stall}}$, target mean score $S_{\mathrm{tar}}$
\Ensure Best-so-far policy under the stored mean-score estimates
\State $(\bar{S}_0,\tau_0)\gets\operatorname{Evaluate}_{R}(\pi_0)$; set $t\gets0$ and $c_{\mathrm{stall}}\gets0$
\While{$t<T\land \bar{S}_t<S_{\mathrm{tar}}\land c_{\mathrm{stall}}<C_{\mathrm{stall}}$}
    \State $\{\pi_{t,j}\}_{j=1}^{k_t}\gets\operatorname{Revise}_{t}(\pi_t;\bar{S}_t,\tau_t)$
    \ForAll{$j=1,\ldots,k_t$ in parallel}
        \State Attempt $\operatorname{Evaluate}_{R}(\pi_{t,j})$; after a failure, repair and restart up to $N_{\mathrm{repair}}$ times
        \If{evaluation completes}
            \State Store $(\bar{S}_{t,j},\tau_{t,j})$
        \EndIf
    \EndFor
    \State Retain the incumbent by default: $(\pi_{t+1},\bar{S}_{t+1},\tau_{t+1})\gets(\pi_t,\bar{S}_t,\tau_t)$
    \State $c_{\mathrm{stall}}\gets c_{\mathrm{stall}}+1$
    \State Select $j_t^{\star}\in\arg\max_{j\in[k_t]}\bar{S}_{t,j}$
    \If{$\bar{S}_{t,j_t^{\star}}>\bar{S}_t$}
        \State Promote candidate $j_t^{\star}$: $(\pi_{t+1},\bar{S}_{t+1},\tau_{t+1})\gets(\pi_{t,j_t^{\star}},\bar{S}_{t,j_t^{\star}},\tau_{t,j_t^{\star}})$
        \State $c_{\mathrm{stall}}\gets0$
    \EndIf
    \State $t\gets t+1$
\EndWhile
\State \Return $\pi_t$
\end{algorithmic}
\end{algorithm}

%% file: sections/experiments.tex
\vspace{-2mm}
\section{Experiments}
\label{sec:experiments}

To evaluate the proposed framework in a complex simulation environment, we design the experiments around three research questions:
\begin{itemize}[leftmargin=7mm,itemsep=1mm, topsep=0em]
    \item \textit{RQ1:} Can the proposed framework outperform representative mathematical-programming, rule-based, and black-box simulation-based optimization baselines in the case-study simulation?
    \item \textit{RQ2:} Can LLM agents use simulation traces to identify operational bottlenecks and make targeted policy modifications?
    \item \textit{RQ3:} How well do the optimized policies perform under changed simulation settings, with and without re-optimization?
\end{itemize}

\subsection{Experimental Settings}
\label{subsec:exp_settings}

\textbf{Simulation configuration.}
The default setting uses a simulation horizon of $T_{\mathrm{sim}}=500$ minutes and a deterministic dynamic order-arrival setting $\Theta_{\mathrm{dyn}}$ in which orders arrive every 10 minutes. The case contains three production lines, shared raw-material and finished-goods warehouses, AGVs, conveyors, buffers, quality inspection, charging, and stochastic product-quality changes. Detailed parameter values, including order distributions, processing times, buffer capacities, and charging parameters, are provided in \Cref{app:default_parameters}.

\textbf{Initial policy.}
The initial policy uses a rule-based structure with three decision layers: production-line selection, transportation-task ranking, and AGV assignment. It ranks tasks mainly by order priority and remaining processing time, then assigns available AGVs. The initial policy does not perform trace-based parameter estimation, but it includes an estimator template that the LLM agents may extend.

\textbf{Baselines.}
The baselines are representative references for mathematical-programming, rule-based, and black-box simulation-based optimization methods. The mathematical programming baseline uses a rolling mixed-integer linear programming (MILP) model for AGV transportation tasks. Because the full simulation includes dynamic order arrivals, quality changes, rework, charging, and blocking, the MILP models a tractable subset of the decision process and re-optimizes when new tasks or rework events appear. The rule-based heuristic baseline is organized as combinations of production-line selection rules, task-ranking rules, and AGV-assignment rules; 135 combinations are evaluated. The conventional metaheuristic baselines include GA, DE, and PSO over the heuristic-combination space. None of the baseline decision spaces includes charging control: the rule combinations cover the three layers above, and the MILP excludes battery and charging constraints (\Cref{app:baselines}).

These baselines represent common optimization paradigms that are effective for well-defined scheduling problems. However, in an evaluation environment with rich event-driven dynamics and operational constraints, mathematical programming models and fixed-dimensional black-box search may have limited ability to identify the operational causes of poor performance. This setting motivates a trace-aware framework that can use simulation traces to diagnose bottlenecks and improve executable scheduling policies.

\textbf{Evaluation protocol.}
All candidate policies generated by the proposed framework are evaluated by repeated simulation and ranked by the mean total score. Unless otherwise stated, the default simulation horizon is $T_{\mathrm{sim}}=500$ minutes, and each candidate evaluation uses $R=10$ scoring replications plus one diagnostic replay; the replay score is excluded from the mean total score. Candidate evaluations use independent seed sets, and the incumbent retains its stored evaluation rather than being re-evaluated using candidate seeds; promotion therefore compares independently estimated mean total scores, and the search does not adapt to one fixed set of realizations. Reported final scores of the framework are the stored mean total scores of each run's final incumbent; the matched-seed re-evaluation in \Cref{app:baseline_evaluation} provides an independent check for the selected policy. The complete optimization procedure is independently repeated five times for each LLM backbone.
The experiments use four LLM backbones: Gemini-3.1-Pro, GPT-5.5, GPT-5.4-mini, and GLM-5. The frontier models are used to probe the performance attainable by the framework in this case, while the lighter models test whether the workflow remains useful under lower-cost models. The selection of $T_{\mathrm{sim}}$ and the remaining runtime parameters of the optimization framework are reported in \Cref{app:horizon_selection,app:runtime_parameters}.

\begin{figure}[t]
    \centering
    \includegraphics[width=.9\linewidth]{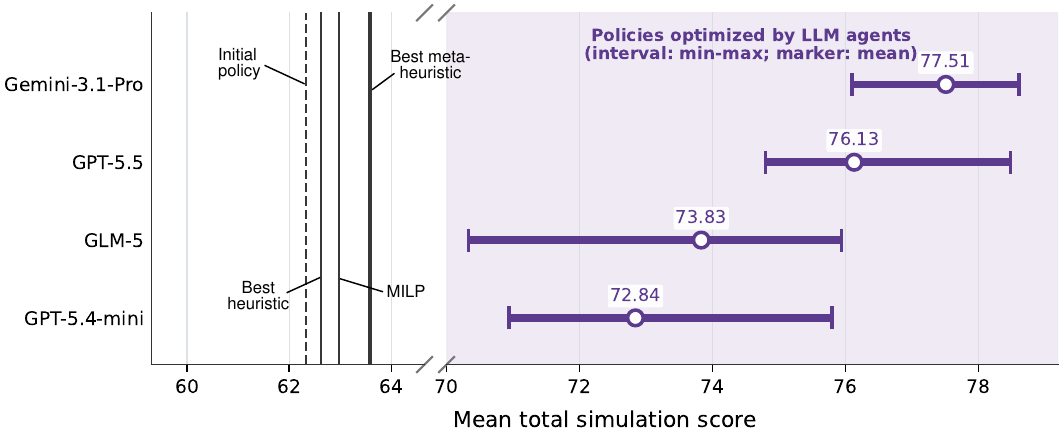}
    \caption{Overall comparison between the proposed framework and representative baselines. Baseline lines denote mean total scores from repeated simulations of fixed policies. For each LLM backbone, the marker and interval denote the mean and the minimum--maximum range of final mean total scores across independent complete optimization runs.}
    \label{fig:baseline_comparison}
\end{figure}

\FloatBarrier
\subsection{Overall Performance Comparison}
\label{subsec:overall_results}

\textit{RQ1} asks whether the proposed framework can outperform conventional optimization baselines in the case-study simulation. \Cref{fig:baseline_comparison} summarizes the mean total scores of the fixed baseline policies and the final mean total scores obtained from independent complete runs of the proposed framework. The best heuristic, MILP, and metaheuristic baselines reach the low-60 score range. In comparison, the LLM-agent framework obtains higher final mean total scores across the tested models, with the best Gemini-3.1-Pro run reaching 78.61 in the default setting.

\begin{figure}[h]
    \centering
    \includegraphics[width=.6\linewidth]{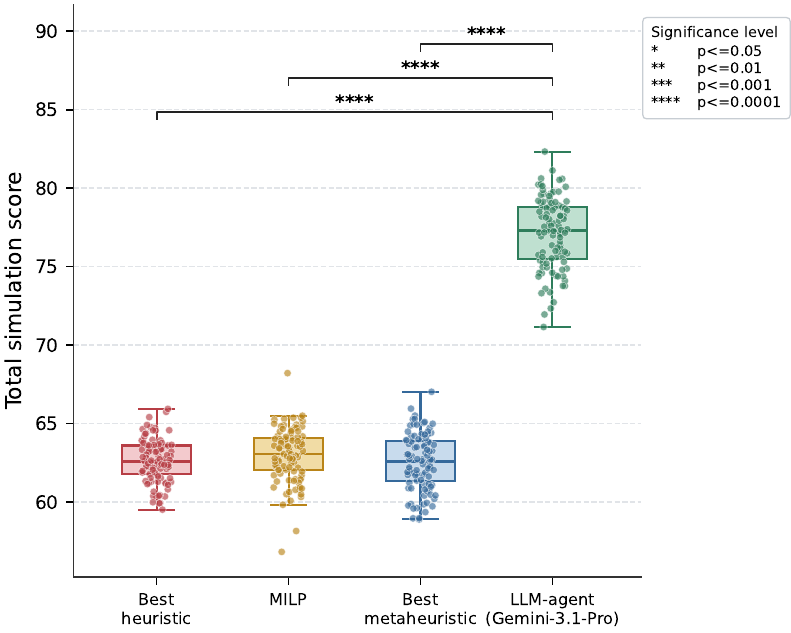}
    \caption{Matched-seed score distributions and hypothesis tests in the default simulation setting.}
    \label{fig:main_boxplot}
\end{figure}

\Cref{fig:main_boxplot} further compares one representative policy from each method family over 100 matched seeds. The LLM-agent group uses the best final Gemini-3.1-Pro policy. The common re-evaluation and statistical testing protocol is reported in \Cref{app:baseline_evaluation}.
Under the common matched-seed protocol, the policy produced by the proposed framework outperforms every representative baseline across all 100 seeds. The differences are statistically significant under two-sided paired Wilcoxon signed-rank tests with Holm correction, providing evidence for the effectiveness of the framework in this case study. Part of this gap reflects the framework's broader decision space rather than trace guidance alone: the baseline policies cannot alter charging behavior (\Cref{subsec:exp_settings}), whereas the optimized policies introduce proactive charging (\Cref{subsec:iterative_improvement}). The ablations in \Cref{subsec:ablation} assess the contribution of trace evidence within this enlarged space.

\begin{figure}[h!]
    \centering
    \includegraphics[width=.6\linewidth]{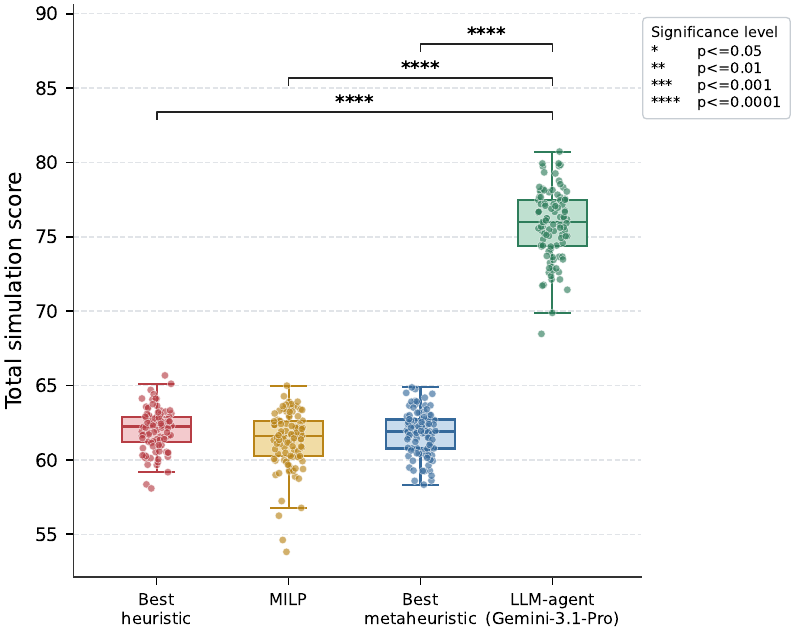}
    \caption{Matched-seed score distributions and hypothesis tests under random faults, without re-optimization.}
    \label{fig:fault_stress_test}
\end{figure}

The final policy obtained in the default setting is also evaluated under random equipment faults without re-optimization, addressing the fixed-policy part of \textit{RQ3}. The comparison keeps the same family-level policy selection as the default-setting boxplot. Each production line has a fault generator; faults may affect workstations, conveyors, or AGVs; inter-fault times follow $U(80,120)$ minutes; and recovery times follow $U(20,60)$ minutes. The matched-seed comparison in \Cref{fig:fault_stress_test}, evaluated using the protocol in \Cref{app:baseline_evaluation}, indicates that the optimized policy retains its advantage when random faults are introduced after optimization.

\begin{figure}[t]
    \centering
    \includegraphics[width=.75\linewidth]{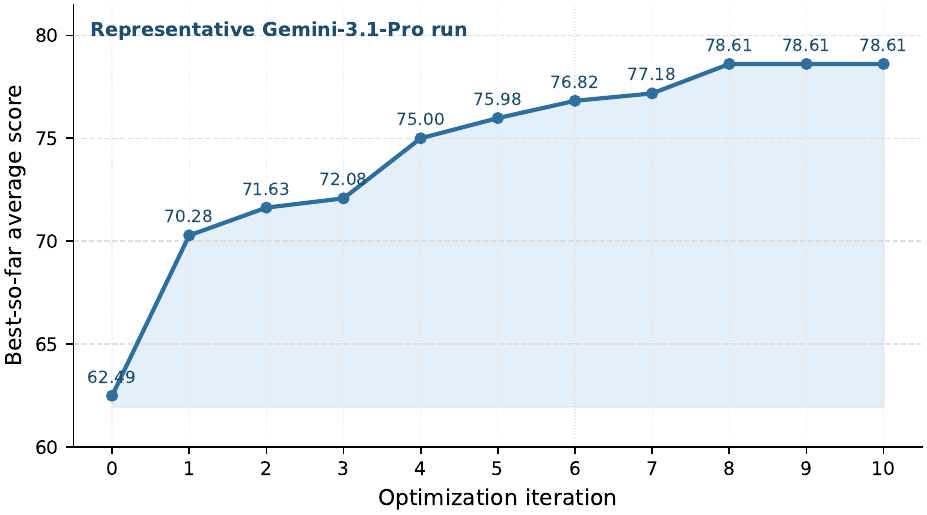}
    \caption{Simulation-trace-guided improvement during a representative optimization run.}
    \label{fig:iteration_analysis}
\end{figure}
\FloatBarrier
\subsection{Simulation-Trace-Guided Iterative Improvement}
\label{subsec:iterative_improvement}

\textit{RQ2} asks whether LLM agents can use simulation traces to identify bottlenecks and make targeted policy modifications. The highest-scoring Gemini-3.1-Pro run provides an illustrative optimization trajectory. Starting from the initial policy, the best-so-far mean total score rises from 62.49 to 78.61 during the 10-iteration run, an improvement of approximately 25.8\%.\footnote{The initial policy corresponds to the DEFAULT rule combination in \Cref{app:heuristic_baselines}.}
This trajectory is summarized in \Cref{fig:iteration_analysis}.

The first trace-based diagnosis concerns inefficient AGV operation. The records show that the AGVs rarely charge proactively and instead enter forced charging after battery depletion. They also indicate substantial empty travel before reaching task origins. The manager agent therefore proposes changes related to active charging, distance-aware assignment, and buffer congestion. The best first-iteration candidate adds a nearby-task preference and sends idle low-battery AGVs to charge proactively, increasing the mean total score from 62.49 to 70.28.
In the next phase, charging behavior improves, but the traces still show limited completed output and delays in rework flow. Some downstream transport tasks have larger effects on these outcomes. The agent therefore increases the priority of tasks that clear downstream buffers or move products closer to completion, while continuing to tune empty-travel penalties and charging thresholds. By the fifth iteration, the best-so-far mean total score reaches 75.98.
Later iterations reveal a different bottleneck. The policy has become too biased toward downstream clearing, which can starve upstream workstations of new material. The agent then adjusts dispatch priority weights and distance penalties to balance front-end feeding with downstream clearing. The best-so-far mean total score reaches 78.61 at iteration 8. The last two iterations make smaller numerical changes to charging and dispatch priorities without improving the incumbent. This trajectory illustrates the role of simulation traces: the LLM links performance changes to operational mechanisms and changes the scheduling logic accordingly.

\FloatBarrier
\subsection{Robustness under Different Simulation Settings}
\label{subsec:robustness}

\begin{figure}[h]
    \centering
    \includegraphics[width=\linewidth]{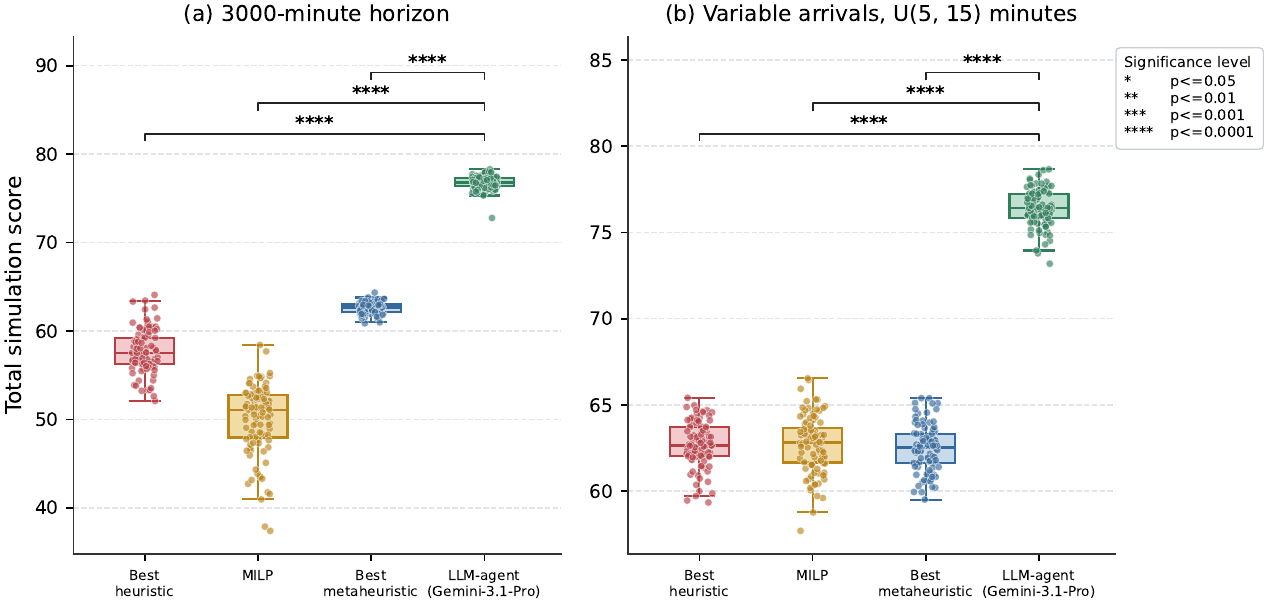}
    \caption{Robustness comparison under changed simulation settings.}
    \label{fig:robustness}
\end{figure}

The second part of \textit{RQ3} asks whether the workflow remains effective when it re-optimizes under changed settings. Unlike the fixed-policy fault test in \Cref{subsec:overall_results}, we conduct a separate optimization under each modified configuration. We consider two changes: increasing the horizon to $T_{\mathrm{sim}}=3000$ minutes and replacing the fixed 10-minute order interarrival time with $U(5,15)$ minutes. For each configuration, the comparison uses the best policy obtained under that configuration for every method family, with Gemini-3.1-Pro used for the proposed framework. The numerical results appear in \Cref{tab:robustness}, and \Cref{fig:robustness} gives the matched-seed comparisons under the protocol in \Cref{app:baseline_evaluation}.

The results suggest a degree of robustness across the tested settings. Simulation-trace-guided policy revision remains effective under a longer $T_{\mathrm{sim}}$ and moderate variability in $\Theta_{\mathrm{dyn}}$.

\begin{table}[t]
\centering
\caption{Performance under changed simulation settings.}
\label{tab:robustness}
\begin{minipage}{0.78\linewidth}
\begin{tabular*}{\linewidth}{@{\extracolsep{\fill}}llc@{}}
\toprule
Setting & Method & Mean total score \\
\midrule
$T_{\mathrm{sim}}=3000$ min & MILP & 50.03 \\
$T_{\mathrm{sim}}=3000$ min & Best heuristic & 57.67 \\
$T_{\mathrm{sim}}=3000$ min & GA & 63.06 \\
$T_{\mathrm{sim}}=3000$ min & DE & 62.91 \\
$T_{\mathrm{sim}}=3000$ min & PSO & 62.96 \\
$T_{\mathrm{sim}}=3000$ min & Proposed framework & 74.16 (72.37, 78.16)\textsuperscript{a} \\
\midrule
Changed $\Theta_{\mathrm{dyn}}$ & MILP & 62.71 \\
Changed $\Theta_{\mathrm{dyn}}$ & Best heuristic & 62.72 \\
Changed $\Theta_{\mathrm{dyn}}$ & GA & 63.42 \\
Changed $\Theta_{\mathrm{dyn}}$ & DE & 63.15 \\
Changed $\Theta_{\mathrm{dyn}}$ & PSO & 62.82 \\
Changed $\Theta_{\mathrm{dyn}}$ & Proposed framework & 76.34 (74.71, 78.97)\textsuperscript{a} \\
\bottomrule
\end{tabular*}
\par\vspace{0.35em}
\raggedright
\footnotesize
\textsuperscript{a}\,For the proposed framework, values in parentheses give the minimum and maximum over five complete optimization runs.
\end{minipage}
\end{table}

\FloatBarrier
\subsection{Ablation Analysis}
\label{subsec:ablation}

The ablation study examines two design choices: parallel candidate generation and access to the trace database. The first ablation allows only one candidate plan per iteration. The second disables the trace database and database-dependent parameter estimation, leaving the agents with only aggregate KPI summaries and policy code. This configuration removes both the manager's trace-based diagnostic evidence and the policies' data-driven parameter estimation; the measured effect reflects their combined contribution. The results are shown in \Cref{tab:ablation}.

The single-candidate configuration retains the potential to find competitive policies: its best Gemini-3.1-Pro run reaches 78.56, close to 78.61 under the original configuration. Its stability, however, declines markedly. The minimum score falls from 76.10 to 62.36 for Gemini-3.1-Pro and from 70.94 to 62.29 for GPT-5.4-mini. The average number of iterations also decreases from 9.2 to 5.8 and from 8.0 to 4.6, respectively. With only one revision direction per iteration, an unproductive candidate is more likely to cause a no-improvement iteration, making the search more susceptible to early stopping.

\begin{table}[H]
\centering
\caption{Ablation results for candidate parallelism and trace-database access.}
\label{tab:ablation}
\begin{tabular}{llcccc}
\toprule
Model & Configuration & Mean & Min. & Max. & Avg. iterations \\
\midrule
Gemini-3.1-Pro & Original & 77.51 & 76.10 & 78.61 & 9.2 \\
Gemini-3.1-Pro & Single candidate & 73.65 & 62.36 & 78.56 & 5.8 \\
Gemini-3.1-Pro & No database & 76.16 & 72.27 & 77.61 & 6.4 \\
GPT-5.4-mini & Original & 72.84 & 70.94 & 75.80 & 8.0 \\
GPT-5.4-mini & Single candidate & 66.22 & 62.29 & 72.07 & 4.6 \\
GPT-5.4-mini & No database & 69.28 & 66.26 & 71.84 & 8.0 \\
\bottomrule
\end{tabular}
\end{table}

Disabling the trace database reduces the mean performance of both models: from 77.51 to 76.16 for Gemini-3.1-Pro and from 72.84 to 69.28 for GPT-5.4-mini, although the across-run ranges overlap for Gemini-3.1-Pro over five runs. The consistent direction of the decline suggests that aggregate KPI summaries and policy code alone provide less effective guidance than when simulation-trace evidence is available. The larger reduction for GPT-5.4-mini further suggests that trace evidence is particularly useful for grounding diagnosis by the lighter model.

%% file: sections/discussions.tex
\FloatBarrier
\section{Discussion}
\label{sec:discussion}

\textbf{Why simulation traces matter.}
The mean score supports candidate selection but provides limited diagnostic detail. Simulation traces connect poor outcomes to mechanisms such as transportation delay, buffer blocking, poor charging timing, quality rework, and delayed urgent orders. In the case study, AGV state records, buffer states, and time-varying metric records allow the agent to identify forced charging, empty travel, downstream clearing, and upstream starvation. This trace-based diagnostic channel distinguishes the proposed framework from scalar-feedback SBO.

\textbf{Relationship to metaheuristic optimization.}
The framework retains the metaheuristic search pattern of generating candidates, evaluating them, selecting better policies, and iterating. Its variation operator acts on executable scheduling code. The LLM can add, remove, or restructure policy logic based on trace evidence, extending the search beyond fixed-dimensional vectors, permutations, rule combinations, and encoded schedules. In the case study, this includes decisions, such as proactive charging, that the baseline rule spaces do not expose. This flexibility also increases the need for execution validation.

\textbf{Between-batch policy revision versus within-run LLM control.}
The framework invokes the LLM between evaluation batches, while a fixed executable policy makes decisions within each simulation run. This separation allows revised policies to be inspected, versioned, and tested before deployment. Within-run LLM control would instead require repeated inference and make runtime behavior more sensitive to inference latency and output variability.

\textbf{General applicability and boundary conditions.} 
The framework is intended for simulation-based optimization tasks in which decision logic can be represented as executable code, candidate policies can be evaluated repeatedly, process-level traces are available, and generated policies can be executed and checked automatically. Although these conditions may hold in manufacturing, warehouse logistics, port operations, intra-hospital logistics, and other discrete-event systems, the empirical evidence in this study comes from dynamic production and AGV scheduling. Operational use would additionally require a verified, validated, and calibrated simulator whose outputs are sufficiently reliable to support policy evaluation.

\textbf{Limitations.} 
Three aspects of the work should be considered when interpreting the results:

\begin{itemize}[leftmargin=7mm,itemsep=1mm, topsep=0em]
\item \textit{Use of optimization history.}
The framework retains previous policies, feedback, and evaluation outcomes, but provides them mainly as contextual records. It does not convert this information into a structured memory of successful revisions, unsuccessful attempts, and their performance effects. The current experiments therefore do not establish how systematically reusing prior search experience would affect optimization.

\item \textit{Incumbent-centered search.} 
The best-so-far mechanism provides a simple and auditable sequence of policy revisions, with every promoted candidate subjected to execution checks and simulation evaluation. However, the search maintains a single incumbent rather than preserving several promising policy branches, which limits diversity when improvement slows.

\item \textit{Stochastic evaluation and trace selection.} Repeated replications reduce dependence on a single simulation realization, and the matched-seed re-evaluation independently checks the reported final comparisons. Candidate promotion nevertheless relies on finite-sample estimates obtained from independent seed sets. Moreover, selecting the lowest-scoring replication for diagnosis deliberately emphasizes observed failure modes, but that trace may not represent typical operating conditions. The current experiments do not isolate the effects of alternative promotion criteria or trace-selection strategies.\end{itemize}

%% file: sections/conclusion.tex
\section{Conclusion}
\label{sec:conclusion}

This paper proposed a framework for LLM-guided heuristic design from simulation traces. Repeated simulation provides a mean score for selection, while a selected diagnostic trace supports policy revision. LLM agents use this evidence to improve executable scheduling policies through code revision between evaluation batches, execution validation, repeated simulation, and best-so-far selection.

In the dynamic production and AGV scheduling case study, the framework improves an initial rule-based policy and outperforms representative mathematical programming, heuristic, and conventional metaheuristic baselines. The iterative analysis and ablation results are consistent with benefits from trace-database access and multi-candidate generation. The results support the value of simulation-trace-guided policy optimization in this simulation setting.

Future work should strengthen the summarization and reuse of historical optimization experience, improve the search mechanism by retaining and revisiting multiple promising policy branches, and test the framework in additional industrial systems such as warehouse scheduling and port logistics. The optimization target can also be extended beyond scheduling rules to higher-level decisions such as resource configuration and buffer capacity.

%% file: sections/appendix.tex
\newpage

\newcolumntype{M}[1]{>{\raggedright\arraybackslash}m{#1}}
\renewcommand{\arraystretch}{1.12}

\section{Simulation Environment and Configuration}
\label[appendix]{app:sim_config}

\subsection{Simulation Case and Product Flows}
\label[appendix]{app:product_flows}

The simulation case contains three production lines. Each line has Workstations A--C, Conveyors AB, BC, and CQ, CQ buffers, a Quality check station, charging access, and two AGVs that serve products on that line. The facilities labeled Raw materials warehouse and Finished goods warehouse are shared across the three lines. Product flows differ by product type, and only part of each route requires AGV transportation. \Cref{tab:product_routes} summarizes the product routes used in the experiments.

\begin{table}[H]
\centering
\caption{Product routes and transportation modes in the simulation case.}
\label{tab:product_routes}
\begin{tabular}{M{0.14\linewidth}M{0.55\linewidth}M{0.2\linewidth}}
\toprule
Product type & Route segment & Transportation mode \\
\midrule
\multirow{6}{*}{A/B} & Raw materials warehouse to Workstation A & AGV \\
 & Workstation A to Workstation B & Conveyor \\
 & Workstation B to Workstation C & Conveyor \\
 & Workstation C to Quality check station & Conveyor \\
 & Quality check station to Workstation C for rework & AGV \\
 & Quality check station to Finished goods warehouse & AGV \\
\midrule
\multirow{9}{*}{C} & Raw materials warehouse to Workstation A & AGV \\
 & Workstation A to Workstation B & Conveyor \\
 & Workstation B to Workstation C & Conveyor \\
 & Workstation C to Conveyor CQ upper/lower buffer & Conveyor \\
 & Conveyor CQ upper/lower buffer to Workstation B & AGV \\
 & Workstation B to Workstation C & Conveyor \\
 & Workstation C to Quality check station & Conveyor \\
 & Quality check station to Workstation C for rework & AGV \\
 & Quality check station to Finished goods warehouse & AGV \\
\bottomrule
\end{tabular}
\end{table}

\FloatBarrier
\subsection{Event-Level Trace Recording}
\label[appendix]{app:trace_recording}

The simulator creates an event-level trace record whenever a DES event changes the state of an entity or performance indicator. These records are later queried and summarized into diagnostic signals for the LLM agents. \Cref{tab:trace_tables} lists the main trace tables and the events that trigger new records.

\begin{table}[t]
\centering
\caption{Event-level trace records exposed by the simulator.}
\label{tab:trace_tables}
\begin{tabular}{M{0.18\linewidth}M{0.24\linewidth}M{0.48\linewidth}}
\toprule
Trace table & Meaning & Events that trigger records \\
\midrule
rawmaterial & Raw materials warehouse state & Simulation start; new raw materials generated by orders; AGV pickup from the Raw materials warehouse \\
warehouse & Finished goods warehouse state & Simulation start; AGV delivery of finished products to the Finished goods warehouse \\
order & Order information & Simulation start; new order generation \\
station & Workstation state & Simulation start; product entering the input buffer; processing start and completion; product transfer to downstream equipment; blocking and unblocking caused by downstream availability; fault occurrence and recovery \\
qualitycheck & Quality check station state & Simulation start; product entering the input buffer; inspection start and completion; pass, rework, or scrap outcome; product entering the output buffer; AGV pickup; blocking and unblocking caused by a full output buffer \\
conveyor & Conveyor state & Simulation start; product entering a conveyor; product transfer to downstream equipment; blocking or unblocking caused by downstream unavailability or full buffers \\
agv & AGV state and tasks & Simulation start; AGV receiving movement task; movement completion; loading and unloading; battery consumption; low-battery warning; active or forced charging; charge completion; fault occurrence and recovery \\
fault & Fault events & Simulation start; fault occurrence; fault recovery \\
kpi & Dynamic KPI values & Simulation start; new order generation; product or order completion; inspection completion; workstation processing start; conveyor transportation start; AGV movement start; AGV charging; simulation end \\
response & External command feedback & Simulation start; AGV commands issued by the external scheduling policy \\
\bottomrule
\end{tabular}
\end{table}

\FloatBarrier
\subsection{Performance Indicators and Score Aggregation}
\label[appendix]{app:metrics}

The simulator first converts eight performance metrics to normalized scores in $[0,100]$. Each normalized score is multiplied by the weight in \Cref{tab:score_indicators}, and the weighted values are summed to obtain the total score $S_r(\pi)$. The weights sum to 100\%, so $S_r(\pi)$ also lies in $[0,100]$. For reporting and diagnostic context, the metrics are grouped into three dimensions. The weights within the production-efficiency, quality-and-cost, and AGV-efficiency groups sum to 40\%, 30\%, and 30\%, respectively.

\begin{table}[H]
\centering
\caption{Performance dimensions, metrics, and weights used to compute the total simulation score.}
\label{tab:score_indicators}
\begin{tabular}{M{0.2\linewidth}M{0.27\linewidth}M{0.42\linewidth}}
\toprule
Performance dimension (aggregate weight) & Metric (weight in total score) & Meaning \\
\midrule
\multirow{3}{*}{\makecell[l]{Production\\efficiency\\(40\%)}} & On-time order completion (16\%) & Percentage of orders completed before their due dates \\
 & Weighted production cycle (16\%) & Average ratio between actual and theoretical production time for qualified products, penalized by product completion rate \\
 & Equipment utilization (8\%) & Average percentage of working time over available time across production equipment \\
\midrule
\multirow{2}{*}{\makecell[l]{Quality and cost\\(30\%)}} & Quality pass rate (12\%) & Percentage of qualified products among all produced products \\
 & Cost-benefit ratio (18\%) & Ratio between baseline production cost and actual cost; actual cost includes material, energy, repair, and scrap costs \\
\midrule
\multirow{3}{*}{\makecell[l]{AGV efficiency\\(30\%)}} & Charging-strategy efficiency (9\%) & Percentage of active charging events among all charging events, including active and forced charging \\
 & AGV energy efficiency (12\%) & Number of transportation tasks completed per unit charging time \\
 & AGV utilization (9\%) & Average percentage of transportation time over AGV available time, excluding fault and charging time \\
\bottomrule
\end{tabular}
\end{table}

\FloatBarrier
\subsection{Default Simulation Parameters}
\label[appendix]{app:default_parameters}

\Cref{tab:resource_parameters,tab:order_parameters,tab:quality_parameters,tab:agv_motion_parameters} report the main default parameters used in the simulation experiments. The three production lines share most parameters; the small number of line-specific differences among the displayed parameters are noted below \Cref{tab:resource_parameters}. Line-specific equipment follows the same longitudinal arrangement, with vertical coordinates shifted across the three lines.

Products with a quality score below 80 at their first inspection are sent to rework, whereas products below 60 are scrapped. A reworked product is also scrapped if its score remains below 80. For AGV motion, travel time follows an acceleration--constant-speed--deceleration profile. When a route requires a turn, the AGV decelerates to rest before accelerating again.

\begin{longtable}{M{0.20\linewidth}M{0.29\linewidth}M{0.41\linewidth}}
\caption{Facility and resource parameters in the default simulation setting.}
\label{tab:resource_parameters}\\
\toprule
Entity & Parameter & Value \\
\midrule
\endfirsthead
\caption[]{Facility and resource parameters in the default simulation setting (continued).}\\
\toprule
Entity & Parameter & Value \\
\midrule
\endhead
\multicolumn{3}{r}{\footnotesize Continued on the next page}\\
\endfoot
\bottomrule
\endlastfoot
\multirow{2}{*}{\makecell[l]{Raw materials\\warehouse}}
 & Position (meter) & $[0,20]$ \\*
 & AGV interaction point & P0 \\
\midrule
\multirow{2}{*}{\makecell[l]{Finished goods\\warehouse}}
 & Position (meter) & $[95,20]$ \\*
 & AGV interaction point & P9 \\
\midrule
\multirow{4}{*}{\makecell[l]{Workstation A}}
 & Position (meter) & $[15,20]$ \\*
 & Input buffer capacity & 2\textsuperscript{c} \\*
 & Processing time (minute) & A: 5; B: 5; C: 5 \\*
 & AGV interaction point & P1 \\
\midrule
\multirow{4}{*}{\makecell[l]{Workstation B}}
 & Position (meter) & $[35,20]$ \\*
 & Input buffer capacity & 2\textsuperscript{c} \\*
 & Processing time (minute) & A: 5; B: 5; C: 5 \\*
 & AGV interaction point & P3 \\
\midrule
\multirow{4}{*}{\makecell[l]{Workstation C}}
 & Position (meter) & $[55,20]$ \\*
 & Input buffer capacity & 2\textsuperscript{c} \\*
 & Processing time (minute) & A: 5; B: 5; C: 5 \\*
 & AGV interaction point & P5 \\
\midrule
\multirow{4}{*}{\makecell[l]{Conveyor AB}}
 & Position (meter) & $[25,20]$ \\*
 & Capacity & 3 \\*
 & Transfer time (minute) & 5 \\*
 & AGV interaction point & P2 \\
\midrule
\multirow{4}{*}{\makecell[l]{Conveyor BC}}
 & Position (meter) & $[45,20]$ \\*
 & Capacity & 3 \\*
 & Transfer time (minute) & 5 \\*
 & AGV interaction point & P4 \\
\midrule
\multirow{5}{*}{\makecell[l]{Conveyor CQ\textsuperscript{a}}}
 & Position (meter) & $[65,20]$ \\*
 & Main-conveyor capacity & 4 \\*
 & Upper/lower-conveyor capacity & 2 \\*
 & Transfer time (minute) & 5 \\*
 & AGV interaction point & P6 \\
\midrule
\multirow{5}{*}{\makecell[l]{Quality check\\station}}
 & Position (meter) & $[75,20]$ \\*
 & Input buffer capacity & 1\textsuperscript{d} \\*
 & Processing time (minute) & P1: 5; P2: 5; P3: 5 \\*
 & Output buffer capacity & 4 \\*
 & AGV interaction points & P7, P8 \\
\midrule
\multirow{5}{*}{\makecell[l]{AGV 1\textsuperscript{b}}}
 & Initial battery & 50\% \\*
 & Load capacity & 2 \\*
 & Charging point & P10 \\*
 & Charging speed (\%/minute) & 3.33 \\*
 & Battery consumption & 0.1\% per meter; 0.5\% per operation \\
\midrule
\multirow{5}{*}{\makecell[l]{AGV 2\textsuperscript{b}}}
 & Initial battery & 50\% \\*
 & Load capacity & 2 \\*
 & Charging point & P10 \\*
 & Charging speed (\%/minute) & 3.33 \\*
 & Battery consumption & 0.5\% per meter; 0.5\% per operation\textsuperscript{e} \\
\end{longtable}
\vspace{-0.8em}
\begin{minipage}{0.94\linewidth}
\footnotesize
\textsuperscript{a}\,Conveyor CQ comprises a main segment from Workstation C to the Quality check station and upper/lower branches with buffers for product C returning to Workstation B.\par
\textsuperscript{b}\,AGVs 1 and 2 serve the Conveyor CQ upper and lower buffers, respectively.\par
\textsuperscript{c}\,The value shown applies to line 1; the corresponding Workstation input-buffer capacity is 3 on lines 2 and 3.\par
\textsuperscript{d}\,The value shown applies to lines 2 and 3; the Quality check station input-buffer capacity is 2 on line 1.\par
\textsuperscript{e}\,The value shown applies to AGV 2 on line 1; AGV 2 consumes 0.1\% per meter on lines 2 and 3. All AGVs consume 0.5\% per operation.
\end{minipage}

\begin{table}[H]
\centering
\caption{Order-generation parameters in the default simulation setting.}
\label{tab:order_parameters}
\begin{tabular}{M{0.42\linewidth}M{0.48\linewidth}}
\toprule
Parameter & Value \\
\midrule
Interarrival time (minute) & 10 \\
Number of products per order & 1: 40\%; 2: 30\%; 3: 20\%; 4: 7\%; 5: 3\% \\
Product type distribution & A: 60\%; B: 30\%; C: 10\% \\
Order priority distribution & Low: 70\%; medium: 25\%; high: 5\% \\
Due-date multiplier\textsuperscript{a} & Low: 3; medium: 2; high: 1.5 \\
\bottomrule
\end{tabular}
\par\vspace{0.35em}
\begin{minipage}{0.90\linewidth}
\footnotesize
\textsuperscript{a}\,Due date $=$ order-generation time $+$ expected processing time $\times$ due-date multiplier. Expected processing times are 160, 200, and 250 minutes for products A, B, and C, respectively.
\end{minipage}
\end{table}

\begin{table}[H]
\centering
\caption{Quality-score dynamics.}
\label{tab:quality_parameters}
\begin{tabular}{M{0.32\linewidth}M{0.58\linewidth}}
\toprule
Event & Quality-score change \\
\midrule
Initial quality score & $U(85,95)$ \\
Handling & 5\% probability of decreasing by $U(1,3)$ \\
Processing & 10\% probability of decreasing by $U(2,5)$ \\
Rework & Restores 70\% of the quality loss caused by processing \\
\bottomrule
\end{tabular}
\end{table}

\begin{table}[H]
\centering
\caption{AGV movement parameters.}
\label{tab:agv_motion_parameters}
\begin{tabular}{M{0.42\linewidth}M{0.48\linewidth}}
\toprule
Parameter & Value \\
\midrule
Loading and unloading time (minute) & 1 per action \\
Travel speed (meter/minute) & 6 \\
Acceleration (meter/minute$^2$) & 1.8 \\
\bottomrule
\end{tabular}
\end{table}

\FloatBarrier
\subsection{Selection of the Simulation Horizon}
\label[appendix]{app:horizon_selection}

\begin{figure}[H]
    \centering
    \includegraphics[width=.92\linewidth]{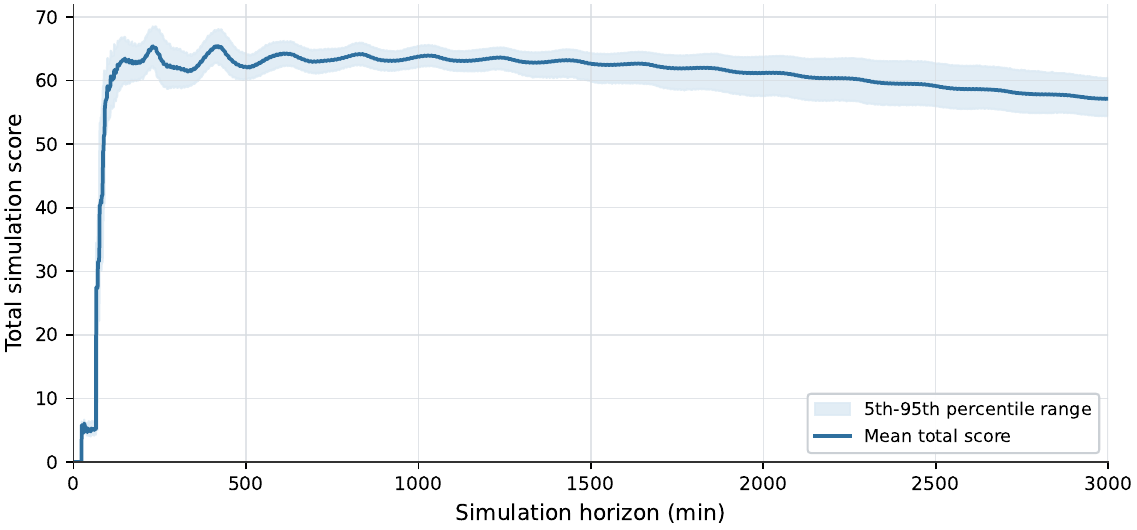}
    \caption{Evolution of the total simulation score under the initial policy. The shaded region shows the 5\%--95\% quantile interval across 1000 replications.}
    \label{fig:horizon_selection}
\end{figure}

To choose the default simulation horizon, the initial policy was simulated for 3000 min over 1000 independent replications. The total score was recorded at every simulated minute, and the mean together with the 5\% and 95\% quantiles were computed across replications. As shown in \Cref{fig:horizon_selection}, the total score increases sharply during the first 100 min, fluctuates between roughly 60 and 65 from 100 to 450 min, and changes only gradually from about 500 min onward, drifting slowly downward over the remainder of the horizon. At 500 min, the 5\%--95\% interval is approximately the mean score plus or minus 2 points. The default experiments therefore use $T_{\mathrm{sim}}=500$ min and reduce residual stochastic variation through repeated replications.

\FloatBarrier
\subsection{Runtime Parameters of the Optimization Framework}
\label[appendix]{app:runtime_parameters}

The main text reports the core evaluation protocol. For completeness, \Cref{tab:runtime_parameters} lists the full runtime parameters used by the LLM-agent optimization framework.

\begin{table}[H]
\centering
\caption{Runtime parameters of the LLM-agent optimization framework.}
\label{tab:runtime_parameters}
\begin{tabular}{M{0.5\linewidth}M{0.36\linewidth}}
\toprule
Parameter & Value \\
\midrule
Maximum manager-agent execution steps & 20 \\
Maximum editing-agent execution steps & 25 \\
Scoring replications per candidate, $R$ & 10 \\
Diagnostic replays per candidate & 1 \\
Maximum optimization iterations, $T$ & 10 \\
No-improvement patience, $C_{\mathrm{stall}}$ & 3 iterations \\
Suggested number of parallel candidates, $k_t$ & 3--4 \\
Maximum repair attempts per failed candidate, $N_{\mathrm{repair}}$ & 3 \\
Target mean score, $S_{\mathrm{tar}}$ & 80 \\
LLM backbones & Gemini-3.1-Pro, GPT-5.5, GPT-5.4-mini, GLM-5 \\
\bottomrule
\end{tabular}
\end{table}

\FloatBarrier
\section{Baseline Methods}
\label[appendix]{app:baselines}

\subsection{Rolling MILP Baseline}
\label[appendix]{app:milp_baseline}

\textbf{Modeling scope and assumptions.}
The mathematical-programming baseline formulates the AGV transportation subproblem as a rolling MILP. Because the full simulator includes dynamic arrivals, quality changes, rework, charging, blocking, and stochastic faults, the MILP uses a tractable abstraction with the following assumptions. First, it explicitly models AGV transportation tasks, while workstation processing, conveyor movement, and quality inspection are controlled by the simulator. Second, although the simulator allows an AGV capacity of two products, the MILP treats each transportation task as moving one product. Third, AGV battery variables and charging constraints are not included in the MILP; feasibility under low battery is handled by the simulator. Fourth, future orders and rework events are unknown under rolling optimization, so each MILP instance only considers current and predictable tasks at the rescheduling time. Fifth, products not yet inspected are assumed to pass inspection when future delivery tasks are predicted.

\textbf{Sets, parameters, and variables.}
Let $t_0$ be the current rescheduling time, $\mathcal{L}$ the set of production lines, $\mathcal{V}$ the set of AGVs, and $\mathcal{V}_{\ell}$ the AGVs assigned to line $\ell$. Let $\mathcal{I}$ be the candidate task set, including raw-material transfer tasks, quality-inspection-to-warehouse tasks, rework tasks, product-C buffer transfer tasks, and predictable future delivery tasks. Tasks are grouped into $\mathcal{G}$, where each group $g$ contains candidate tasks $\mathcal{I}_g$. For raw-material line-selection tasks, $|\mathcal{I}_g|>1$; for other tasks, $|\mathcal{I}_g|=1$.

For each task $i$, $o_i$ and $d_i$ denote its origin and destination, $r_i$ its release time, $\mathcal{V}_i$ its executable AGV set, and $p_i$ its duration. The travel time from location $a$ to location $b$ is denoted by $T^{\mathrm{trav}}(a,b)$. The duration is $p_i=T^{\mathrm{trav}}(o_i,d_i)+2\bar{h}_i$, where $\bar{h}_i$ is the average loading or unloading time over executable AGVs. For each AGV $v$, $q_v$ is its expected starting location and $a_v$ is its expected available time. Let $\mathcal{P}_v=\{(i,h):i,h\in\mathcal{I},\ i<h,\ v\in\mathcal{V}_i\cap\mathcal{V}_h\}$ be the unordered distinct task pairs that AGV $v$ can execute. The decision variables are: $z_i$, indicating whether candidate task $i$ is selected; $x_{i,v}$, indicating whether task $i$ is assigned to AGV $v$; $s_i$ and $c_i$, the planned start and completion times; $C_{\max}$, the maximum task completion time; and $u_{i,h,v}$, indicating whether task $i$ precedes task $h$ on AGV $v$. Both directions $u_{i,h,v}$ and $u_{h,i,v}$ are defined for each $(i,h)\in\mathcal{P}_v$.

\textbf{Constraints and objective.}
The rolling MILP is:
\begin{align}
\sum_{i\in\mathcal{I}_g} z_i &= 1, && \forall g\in\mathcal{G}, \\
\sum_{v\in\mathcal{V}_i} x_{i,v} &= z_i, && \forall i\in\mathcal{I}, \\
s_i &\ge r_i - M(1-z_i), && \forall i\in\mathcal{I}, \\
c_i &\ge s_i + p_i - M(1-z_i), && \forall i\in\mathcal{I}, \\
c_i &\le s_i + p_i + M(1-z_i), && \forall i\in\mathcal{I}, \\
C_{\max} &\ge c_i - M(1-z_i), && \forall i\in\mathcal{I}, \\
s_i &\ge a_v + T^{\mathrm{trav}}(q_v,o_i) - M(1-x_{i,v}), && \forall i\in\mathcal{I}, v\in\mathcal{V}_i, \\
u_{i,h,v} &\le x_{i,v},\quad u_{i,h,v}\le x_{h,v}, && \forall v\in\mathcal{V}, (i,h)\in\mathcal{P}_v, \\
u_{h,i,v} &\le x_{h,v},\quad u_{h,i,v}\le x_{i,v}, && \forall v\in\mathcal{V}, (i,h)\in\mathcal{P}_v, \\
u_{i,h,v}+u_{h,i,v} &\ge x_{i,v}+x_{h,v}-1, && \forall v\in\mathcal{V}, (i,h)\in\mathcal{P}_v, \\
u_{i,h,v}+u_{h,i,v} &\le 1, && \forall v\in\mathcal{V}, (i,h)\in\mathcal{P}_v, \\
s_h &\ge c_i + T^{\mathrm{trav}}(d_i,o_h) - M(1-u_{i,h,v}), && \forall v\in\mathcal{V}, (i,h)\in\mathcal{P}_v, \\
s_i &\ge c_h + T^{\mathrm{trav}}(d_h,o_i) - M(1-u_{h,i,v}), && \forall v\in\mathcal{V}, (i,h)\in\mathcal{P}_v.
\end{align}
The binary variables are $z_i$, $x_{i,v}$, and $u_{i,h,v}$; the timing variables are nonnegative. In the implementation, the big-$M$ value is constructed in the simulator's internal time units as
\begin{equation}
p_{\max}=\max_{i\in\mathcal{I}}p_i,
\qquad
H=t_0+|\mathcal{I}|(p_{\max}+20+3)+60,
\qquad
M=\max\{10^4,2H\}.
\end{equation}
The objective is
\begin{equation}
\min \; C_{\max} + \epsilon \sum_{i\in\mathcal{I}} s_i,
\end{equation}
where $\epsilon=10^{-4}$ is a small tie-breaking coefficient that favors earlier task starts when makespan values are close.

\textbf{Rolling solution procedure and two-stage variant.}
The simulator triggers replanning at initialization and when the number of new orders or rework events increases. To avoid excessive repeated optimization, consecutive calls are separated by at least 8 simulation-time units. Each MILP instance is solved with Gurobi using a 2 s wall-clock limit. The implementation accepts only an optimal solver status; any other termination, including a time limit with an incumbent solution, invokes the fallback procedure. Under this procedure, tasks are ordered by release time and task identifier and assigned greedily to the eligible AGV with the earliest predicted completion time.

The original rolling model solves a single MILP over all candidate tasks. The two-stage variant first assigns each raw-material product to the line $\ell$ minimizing
\begin{equation}
\min_{v\in\mathcal{V}_i}
\left[a_v+T^{\mathrm{trav}}(q_v,o_i)\right]
+T^{\mathrm{trav}}(o_i,d_i)+3n_{\ell},
\end{equation}
where $n_{\ell}$ is the number of raw-material products already assigned to line $\ell$ during that stage. It then solves a separate MILP for each production line. Separating line assignment from within-line task scheduling reduces the size and coupling of each optimization problem; in the experiments, it also reduces fallback frequency. If a per-line model is rejected, the same release-time-ordered greedy AGV assignment is used. During execution, the dispatcher follows planned start times when possible; if no planned task is ready, it may dispatch another currently ready task. The MILP times therefore guide the execution queue rather than imposing hard execution times. As shown in \Cref{tab:milp_variant_results}, the two-stage variant achieves a higher mean score in the default setting. It is therefore used for evaluation throughout the paper. Unless the original and two-stage variants are explicitly distinguished in this appendix, every result labeled ``MILP'' refers to the two-stage variant.

\begin{table}[t]
\centering
\caption{Results of the original and two-stage rolling MILP variants in the default setting. The mean, minimum, and maximum are computed over repeated simulations of each fixed policy.}
\label{tab:milp_variant_results}
\begin{tabular}{lccc}
\toprule
MILP variant & Mean & Min. & Max. \\
\midrule
Original rolling model & 61.19 & 54.56 & 64.58 \\
Two-stage rolling model & 62.97 & 56.82 & 68.21 \\
\bottomrule
\end{tabular}
\end{table}

\FloatBarrier
\subsection{Rule-Based Heuristic Baselines}
\label[appendix]{app:heuristic_baselines}

The rule-based baselines combine three decision layers: production-line selection, transportation-task ranking, and AGV assignment. The candidate rules are listed in \Cref{tab:heuristic_rules}. Including the default initial-rule component at each layer, the experiment evaluates $5\times9\times3=135$ combinations. Each combination is evaluated for 500 simulated minutes over 100 replications under the same random seeds. \Cref{tab:heuristic_top10} reports the ten best combinations.

\begin{table}[H]
\centering
\small
\renewcommand{\arraystretch}{1.0}
\caption{Rule components used in the heuristic baselines.}
\label{tab:heuristic_rules}
\begin{tabular}{M{0.18\linewidth}M{0.22\linewidth}M{0.48\linewidth}}
\toprule
Decision layer & Rule & Description \\
\midrule
\multirow{5}{*}{\makecell[l]{Line\\selection}} & DEFAULT & Initial line-selection logic used in the initial policy \\
 & SQ & Select the line with the smallest queue size \\
 & LWT & Select the line with the least remaining workload \\
 & MET & Select the line with the minimum theoretical execution time for the product \\
 & RS & Randomly select a line \\
\midrule
\multirow{9}{*}{\makecell[l]{Task\\ranking}} & DEFAULT & Initial task-ranking logic used in the initial policy \\
 & SPT & Prefer the task with the shortest next processing time \\
 & LWKR & Prefer the task with the least remaining work \\
 & LOPNR & Prefer the task with the least number of remaining operations \\
 & EDD & Prefer the task with the earliest due date \\
 & CR & Prefer the task by critical ratio, defined by remaining time to due date over remaining work \\
 & MS & Prefer the task with minimum slack \\
 & FIFO & Prefer tasks generated earlier \\
 & RS & Randomly select a task \\
\midrule
\multirow{3}{*}{\makecell[l]{AGV\\assignment}} & DEFAULT & Initial AGV-assignment logic used in the initial policy \\
 & NVF & Select the available AGV nearest to the task origin \\
 & RS & Randomly select an available AGV \\
\bottomrule
\end{tabular}
\end{table}

\begin{table}[t]
\centering
\caption{Top ten rule-based heuristic combinations in the default setting.}
\label{tab:heuristic_top10}
\begin{tabular}{lllccc}
\toprule
Line selection & Task ranking & AGV assignment & Mean & Min. & Max. \\
\midrule
DEFAULT & LWKR & NVF & 62.62 & 59.52 & 65.93 \\
DEFAULT & LWKR & DEFAULT & 62.60 & 59.93 & 65.93 \\
DEFAULT & LWKR & RS & 62.37 & 58.53 & 65.49 \\
DEFAULT & DEFAULT & NVF & 62.36 & 59.69 & 64.64 \\
DEFAULT & DEFAULT & DEFAULT & 62.33 & 59.69 & 64.64 \\
DEFAULT & DEFAULT & RS & 62.23 & 58.40 & 65.16 \\
SQ & LWKR & RS & 62.15 & 58.65 & 65.18 \\
MET & LWKR & NVF & 62.07 & 59.49 & 65.08 \\
MET & LWKR & DEFAULT & 62.05 & 59.41 & 64.88 \\
SQ & LWKR & NVF & 62.01 & 58.58 & 64.46 \\
\bottomrule
\end{tabular}
\end{table}

\FloatBarrier
\subsection{Conventional Metaheuristic Baselines}
\label[appendix]{app:metaheuristic_baselines}

The conventional metaheuristic baselines search over weighted combinations of the non-random heuristic rules in \Cref{tab:heuristic_rules}. Let $d\in\{L,T,V\}$ denote the line-selection, task-ranking, and AGV-assignment layers, respectively, and let $H_d$ be the set of rules available at layer $d$. A candidate is the 14-dimensional vector
\begin{equation}
\boldsymbol{x}
=\left(\boldsymbol{w}^{L},\boldsymbol{w}^{T},\boldsymbol{w}^{V}\right)
\in \Delta_{3}\times\Delta_{7}\times\Delta_{1},
\qquad
\Delta_{m}=\left\{\boldsymbol{w}\in\mathbb{R}_{+}^{m+1}:\sum_{h=1}^{m+1}w_h=1\right\},
\end{equation}
where $\boldsymbol{w}^{L}\in\mathbb{R}^{4}$ corresponds to DEFAULT, SQ, LWT, and MET; $\boldsymbol{w}^{T}\in\mathbb{R}^{8}$ corresponds to DEFAULT, SPT, LWKR, LOPNR, EDD, CR, MS, and FIFO; and $\boldsymbol{w}^{V}\in\mathbb{R}^{2}$ corresponds to DEFAULT and NVF. Random-selection rules are excluded from this continuous search space.

At a decision epoch, let $\mathcal{A}_d$ be the feasible alternatives for layer $d$, and let $g_h^{d}(a)$ be the score assigned to alternative $a\in\mathcal{A}_d$ by rule $h$. Because the component rules use different score scales, their outputs are converted to rank scores. If $\operatorname{rank}_h^d(a)\in\{0,\ldots,|\mathcal{A}_d|-1\}$ is the rank of $a$ under rule $h$, with rank 0 denoting the best alternative, then
\begin{equation}
\widetilde{g}_h^{d}(a)=
\begin{cases}
1-\dfrac{\operatorname{rank}_h^d(a)}{|\mathcal{A}_d|-1}, & |\mathcal{A}_d|>1,\\
1, & |\mathcal{A}_d|=1.
\end{cases}
\end{equation}
The combined priority and selected alternative are
\begin{equation}
G_d(a\mid\boldsymbol{w}^{d})
=\sum_{h\in H_d}w_h^{d}\widetilde{g}_h^{d}(a),
\qquad
a_d^{*}\in\arg\max_{a\in\mathcal{A}_d}G_d(a\mid\boldsymbol{w}^{d}).
\end{equation}
Thus, a metaheuristic searches the weights of an interpretable composite dispatch rule rather than directly encoding a complete schedule.

Let $\pi_{\boldsymbol{x}}$ denote the composite dispatch policy induced by candidate vector $\boldsymbol{x}$. Its sample standard deviation over $R$ replications is
\begin{equation}
\widehat{\sigma}_R(\pi_{\boldsymbol{x}})
=\sqrt{
\frac{1}{R-1}\sum_{r=1}^{R}
\left[S_r(\pi_{\boldsymbol{x}})-\bar{S}_R(\pi_{\boldsymbol{x}})\right]^2
}.
\end{equation}
The stability-penalized baseline fitness is
\begin{equation}
\Phi_R(\boldsymbol{x})
=\bar{S}_R(\pi_{\boldsymbol{x}})
-\lambda\widehat{\sigma}_R(\pi_{\boldsymbol{x}}),
\end{equation}
where $\lambda=0.35$. Unlike the proposed framework's mean-score selection rule, this baseline fitness penalizes variation across replications. The initial population consists of all one-hot combinations of non-random heuristic rules, giving $4\times8\times2=64$ individuals. GA, DE, and PSO are each run for 10 generations, and each candidate is evaluated over 20 simulation replications. Hyperparameters are listed in \Cref{tab:metaheuristic_params}.

More specifically, the initial population is
\begin{equation}
\mathcal{P}_0
=\left\{(\boldsymbol{e}_i^{L},\boldsymbol{e}_j^{T},\boldsymbol{e}_k^{V}):
i=1,\ldots,4;\ j=1,\ldots,8;\ k=1,2\right\},
\end{equation}
where $\boldsymbol{e}$ denotes a one-hot vector. This construction embeds every deterministic non-random rule combination in the initial population. After any search operator is applied, each segment is mapped back to its simplex by a repair-and-normalization operator $\operatorname{Repair}_{\Delta}$.

For GA, simulated binary crossover (SBX) produces two offspring from parents $\boldsymbol{x}^{(1)}$ and $\boldsymbol{x}^{(2)}$ as
\begin{equation}
\boldsymbol{y}^{(1,2)}
=\operatorname{Repair}_{\Delta}\!\left(
\frac{1}{2}\left[(1\pm\boldsymbol{\beta})\odot\boldsymbol{x}^{(1)}
+(1\mp\boldsymbol{\beta})\odot\boldsymbol{x}^{(2)}\right]
\right),
\end{equation}
where the elementwise spread factor $\boldsymbol{\beta}$ is sampled according to the SBX distribution index $\eta_c=15$. Polynomial mutation with index $\eta_m=20$ then perturbs selected coordinates before the same segmentwise repair is applied.

For DE/best/1/bin, the mutant and crossover vectors for target $\boldsymbol{x}_{i,g}$ at generation $g$ are
\begin{align}
\boldsymbol{\nu}_{i,g} &= \boldsymbol{x}_{\mathrm{best},g}
+F\left(\boldsymbol{x}_{i_1,g}-\boldsymbol{x}_{i_2,g}\right),\\
\widetilde{u}_{i,g,j} &=
\begin{cases}
\nu_{i,g,j}, & \rho_j\le \mathrm{CR}\ \text{or}\ j=j_{\mathrm{rand}},\\
x_{i,g,j}, & \text{otherwise},
\end{cases}
\end{align}
where $\boldsymbol{x}_{\mathrm{best},g}$ is sampled from the current rank-zero individuals; $i_1$ and $i_2$ are distinct from each other and from target $i$ but are not separately constrained to differ from the selected best-base index; $\rho_j\sim U(0,1)$; and $j_{\mathrm{rand}}$ is a randomly selected coordinate that guarantees at least one donor coordinate. The implementation uses $F=0.5$ and $\mathrm{CR}=0.2$. With probability 0.1 at the trial-vector level, polynomial mutation with distribution index $\eta_m=20$ is then applied. Conditional on mutation, coordinates are selected with the default probability $1/14$, with at least one coordinate selected. Segmentwise simplex repair is applied to the resulting trial vector, and the repaired trial replaces the target when it has higher fitness.

For PSO, particle $i$ is updated by
\begin{align}
\boldsymbol{v}_{i,g+1}
&=\omega\boldsymbol{v}_{i,g}
+c_1\boldsymbol{\rho}_{1,g}\odot(\boldsymbol{p}_{i,g}-\boldsymbol{x}_{i,g})
+c_2\boldsymbol{\rho}_{2,g}\odot(\boldsymbol{x}_{g}^{\mathrm{gbest}}-\boldsymbol{x}_{i,g}),\\
\boldsymbol{x}_{i,g+1}
&=\operatorname{Repair}_{\Delta}\!\left(\boldsymbol{x}_{i,g}+\operatorname{clip}(\boldsymbol{v}_{i,g+1},-v_{\max},v_{\max})\right),
\end{align}
where $\boldsymbol{p}_{i,g}$ and $\boldsymbol{x}_{g}^{\mathrm{gbest}}$ are the personal and global best positions, $\boldsymbol{\rho}_{1,g},\boldsymbol{\rho}_{2,g}\sim U(0,1)^{14}$, $\omega=0.9$, $c_1=c_2=2.0$, and $v_{\max}$ is 20\% of the corresponding coordinate range. These operators differ in how they explore the weight space, but all candidates are evaluated using the baseline fitness $\Phi_R$.

\begin{table}[H]
\centering
\caption{Hyperparameters of the conventional metaheuristic baselines.}
\label{tab:metaheuristic_params}
\begin{tabular}{M{0.2\linewidth}M{0.45\linewidth}M{0.25\linewidth}}
\toprule
Algorithm & Hyperparameter & Value \\
\midrule
\multirow{5}{*}{GA} & SBX crossover distribution index & 15 \\
 & Polynomial mutation distribution index & 20 \\
 & Crossover probability & 0.9 \\
 & Per-variable crossover probability & 0.5 \\
 & Mutation probability & 0.25 \\
\midrule
\multirow{5}{*}{DE} & DE variant & DE/best/1/bin \\
 & Differential weight & 0.5 \\
 & Crossover rate & 0.2 \\
 & Polynomial-mutation distribution index & 20 \\
 & Trial-vector polynomial-mutation probability & 0.1 \\
\midrule
\multirow{5}{*}{PSO} & Inertia weight & 0.9 \\
 & Cognitive learning factor & 2.0 \\
 & Social learning factor & 2.0 \\
 & Maximum velocity ratio & 0.2 \\
 & Initial velocity setting & random \\
\bottomrule
\end{tabular}
\end{table}

\begin{table}[H]
\centering
\caption{Results of the conventional metaheuristic baselines in the default setting. The mean, minimum, and maximum are computed over repeated simulations of the selected policy.}
\label{tab:metaheuristic_results}
\begin{tabular}{lccc}
\toprule
Algorithm & Mean & Min. & Max. \\
\midrule
GA & 63.22 & 61.08 & 66.26 \\
DE & 63.58 & 61.53 & 66.14 \\
PSO & 63.46 & 61.40 & 65.40 \\
\bottomrule
\end{tabular}
\end{table}

\FloatBarrier
\subsection{Common Re-evaluation and Statistical Comparison Protocol}
\label[appendix]{app:baseline_evaluation}

For each setting, one representative policy is frozen from each method family before a separate common re-evaluation: the two-stage MILP, the best rule-based policy, the best conventional metaheuristic policy, and the best final Gemini-3.1-Pro policy. The four policies are evaluated over the same 100 seeds (123--222). Matching seeds provides common exogenous randomness, although policy-dependent event sequences need not produce identical realized trajectories.

For policy $q$, write $S_n^{(q)}=S(\pi_q;\xi_n)$ for its score under matched stochastic input $\xi_n$. For each baseline $b$, the paired difference is $D_n^{(b)}=S_n^{\mathrm{LLM}}-S_n^{(b)}$. We test whether these paired differences are centered at zero using a two-sided Wilcoxon signed-rank test. For the nonzero differences, the absolute values $|D_n^{(b)}|$ are ranked; $W^{+}$ and $W^{-}$ denote the sums of ranks associated with positive and negative differences, respectively, and the two-sided test statistic is $W=\min(W^{+},W^{-})$. The three baseline-comparison $p$-values within each setting are adjusted using the Holm method. In all 12 comparisons, the LLM-agent policy scores higher for every one of the 100 matched seeds. Consequently, $W^{-}=0$ and hence $W=0$. The asymptotic two-sided $p$-value is $3.90\times10^{-18}$, and the Holm-adjusted asymptotic value is $1.17\times10^{-17}$ for each comparison. \Cref{tab:paired_tests} reports the median paired differences as measures of the observed performance gaps.

\begin{table}[H]
\centering
\caption{Matched-seed comparisons between the LLM-agent policy and the representative baselines. Each entry is the median paired score difference $\widetilde{D}^{(b)}=\operatorname{median}_{n}D_n^{(b)}$ over 100 seeds.}
\label{tab:paired_tests}
\begin{tabular}{lccc}
\toprule
Setting & Heuristic & MILP & Metaheuristic \\
\midrule
Default & 14.45 & 14.36 & 15.11 \\
Random faults & 14.25 & 14.57 & 14.02 \\
$T_{\mathrm{sim}}=3000$ min & 19.50 & 25.79 & 14.18 \\
Variable interarrival time & 13.82 & 13.56 & 13.85 \\
\bottomrule
\end{tabular}
\end{table}